# Biases in Expected Goals Models Confound Finishing Ability

Jesse Davis *
jesse.davis@kuleuven.be

Pieter Robberechts *
pieter.robberechts@kuleuven.be

KU Leuven, Dept. of Computer Science; Leuven.AI,
B-3000 Leuven, Belgium

**Abstract.** Expected Goals (xG) has emerged as a popular tool for evaluating finishing skill in soccer analytics. It involves comparing a player's cumulative xG with their actual goal output, where consistent overperformance indicates strong finishing ability. However, the assessment of finishing skill in soccer using xG remains contentious due to players' difficulty in consistently outperforming their cumulative xG. In this paper, we aim to address the limitations and nuances surrounding the evaluation of finishing skill using xG statistics. Specifically, we explore three hypotheses: (1) the deviation between actual and expected goals is an inadequate metric due to the high variance of shot outcomes and limited sample sizes, (2) the inclusion of all shots in cumulative xG calculation may be inappropriate, and (3) xG models contain biases arising from interdependencies in the data that affect skill measurement. We found that sustained overperformance of cumulative xG requires both high shot volumes and exceptional finishing, including all shot types can obscure the finishing ability of proficient strikers, and that there is a persistent bias that makes the actual and expected goals closer for excellent finishers than it really is. Overall, our analysis indicates that we need more nuanced quantitative approaches for investigating a player's finishing ability, which we achieved using a technique from AI fairness to learn an xG model that is calibrated for multiple subgroups of players. As a concrete use case, we show that (1) the standard biased xG model underestimates Messi's GAX by 17% and (2) Messi's GAX is 27% higher than the typical elite high-shot-volume attacker, indicating that Messi is even a more exceptional finisher than people commonly believed.

## 1. Introduction

Advanced metrics based on machine learning models are increasingly being used in sports, providing valuable insights into player performance, team performance and game strategies. Among these metrics, expected value metrics stand out as one of the most common types. They aim to measure the expected outcome of an event by considering various factors related to the current game state and have been proposed for many sports, including American Football (e.g., expected completion percentage for quarterbacks [1] and expected yards after the catch for receivers [2]), basketball (e.g., expected field goal percentage [3]), and ice hockey (e.g., expected goals [4]). In the realm of soccer, a prominent and widely used expected value metric is expected goals (xG) [5]–[8]. This metric is designed to quantify the quality of scoring opportunities. To compute the xG metric, a binary goal / no goal classifier is trained using historical examples of shots, with each shot described by specific features of the game state at the time of the shot.

Expected value metrics serve multiple purposes in sports analytics. Firstly, they help mitigate the effects of limited sample sizes by providing a more smoothed representation of observed outcomes. In soccer, goals are infrequent, and the conversion of shots can be subject to variations caused by factors like deflections off a player or the goalposts. Hence, cumulative xG is a more stable indicator

* Both authors contributed equally.

of a team's ability to generate scoring chances. Secondly, these metrics offer valuable contextual information. For instance, a soccer player may exhibit a low shot conversion rate due to a tendency to take speculative shots.

This paper focuses on a third application of expected value metrics: their utilization as a tool to measure player skill. If the underlying machine learning model does not take the quality of the player(s) involved in a particular play into account, people believe that the expected outcomes can serve as an approximation of how an average player or team would likely perform in a similar situation. Thus, the difference between the actual observed outcomes and the expected outcomes can be interpreted as an indicator of a player's skill.[1] In the context of finishing skill in soccer, this comparison entails examining the difference between the number of goals scored by a player and the cumulative xG of all their shots within a given time frame, resulting in a *goals above expectation* (GAX)[2] measure:

$$GAX = \sum_{i}^{N} (G_i - xG_i) \tag{1}$$

with $N$ the number of shots and $G \in \{0,1\}$ representing a miss or goal, respectively.

However, evaluating finishing skill in soccer remains a contentious topic within the analytics community due to the notable difficulty players face in consistently outperforming their cumulative xG [9]. In this paper, we assume that finishing skill exists and explore why the disparity between a player's actual and expected goals might yet fail to effectively capture it. Therefore, we use a mix of simulation studies and analysis of real data and find that:

1. Players are unlikely to overperform their cumulative xG unless they can pair exceptional finishing with high shot volumes due to limited sample sizes and high variances. Thus, in itself, GAX is a poor metric to measure finishing skill.
2. Including all shots in the cumulative xG total is not appropriate. Finishing skill has multiple facets that should be looked at individually (e.g., headers vs. long-range shots) and some shots do not reflect a player's finishing skill (e.g., deflected shots).
3. Using GAX to measure finishing skill assumes that xG captures the probability that the "average player" would finish a shot. Our analysis shows that this assumption is violated due to an overrepresentation of shots from good finishers in the data. Consequently, xG models exhibit a persistent bias that makes the actual and expected goals closer for excellent finishers than it really is.

This highlights the need for more nuanced quantitative approaches for investigating a player's finishing ability. We borrow ideas from fairness in artificial intelligence literature to better elucidate and mitigate the bias in xG models. Specifically, we propose using the idea of multi-calibration to learn an xG model that is calibrated for multiple subgroups of players. This allows us to explicitly state characteristics of the player group used to derive the GAX metric. As a concrete use case, we consider Messi's finishing skill with respect to subpopulations defined by a player's shooting volume and playing position. This analysis shows that (1) the standard biased xG model underestimates Messi's GAX by 17% and (2) Messi's GAX is 27% higher than the typical elite high-shot-volume attacker.

[1] For example, FBref defines xG as "an estimate of how the average player or team would perform in a similar situation" (see https://fbref.com/en/expected-goals-model-explained/). This definition is used by popular media to justify xG as a tool to evaluate finishing skill (e.g., http://www.bbc.com/sport/football/66806572 and https://www.espn.com/soccer/story/_/id/37577474).

[2] The alternate terms "Goals Above Average (GAA)" and "xG Above Replacement (xGAR)" have also been used for the same metric. Additionally, the GAX is sometimes defined as the ratio G / xG.

## 2. Hypotheses About Finishing Skill

We start from the assumption that finishing skill exists and detail three potential reasons why GAX may not be able to capture this skill. We empirically analyze the hypotheses, primarily using simulation studies.

### Hypothesis 1: Limited sample sizes, high variances and small variations in skill between players make GAX a poor metric for measuring finishing skill.

Overall conversion rates on shots are low and the variance of goals scored is high (Figure 1). In the Big Five European leagues, only 9% of players (with at least one shot) take more than 50 shots in a season and they convert about 11% of these shots. Therefore, GAX provides limited empirical discrimination and stability [10].

One possible remedy would be to aggregate over consecutive seasons to increase the sample size. This is valid if we assume that finishing skill is a static quantity. However, it seems plausible that finishing skill varies over time with players improving when they are younger and declining when they age.

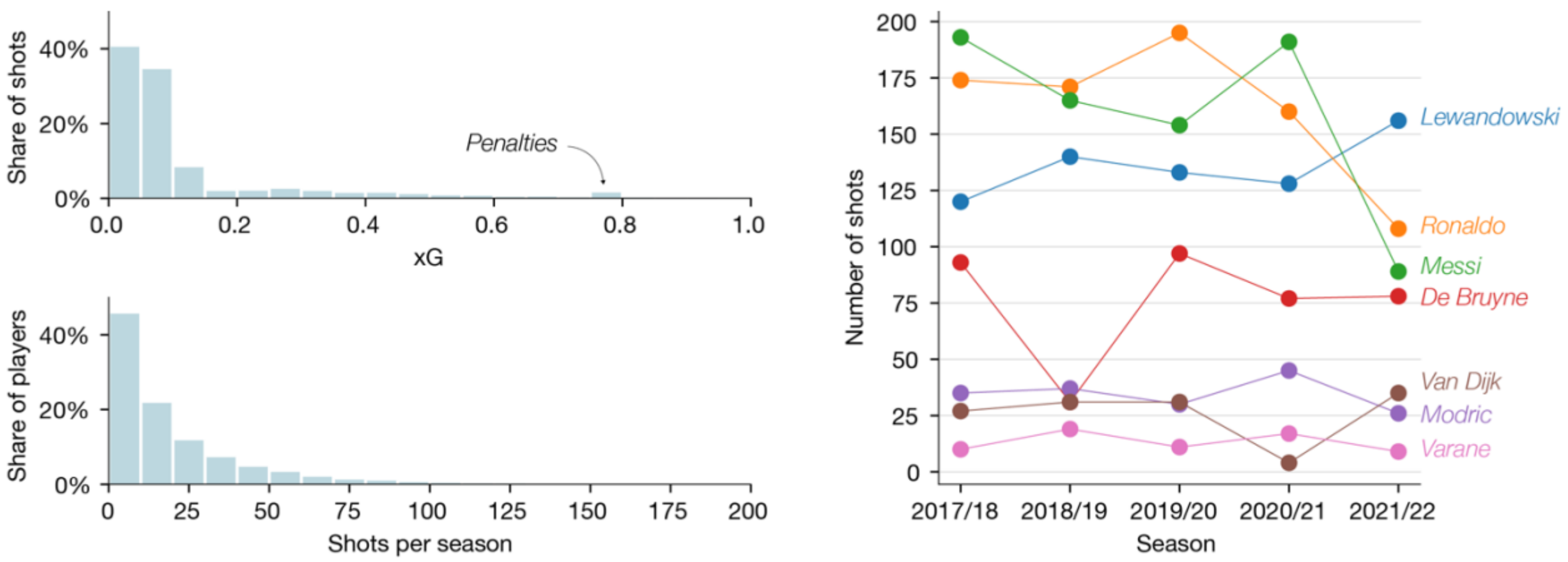

*Figure 1. Distribution of xG values (top left) and the number of shots attempted per season (bottom left and right) in the Big Five European leagues between 2017/18 and 2021/22. Data source: FBref.*

Third, even good finishers may convert their chances with only slightly higher rates than average ones. By all accounts, Messi is an outstanding finisher and high-volume shooter, and he has only outperformed his xG by around 25% over his entire career.[3]

We ran a simulation experiment to assess how many shots are needed in order to be likely to observe a difference in GAX if a player is an $\alpha$% better finisher than an average player. First, we train an xG model using logistic regression on StatsBomb Open Data for the 2015/16 season of the Big Five European leagues, which contains 43,110 open play shots. We consider the following features: the shots' location ($x, y$ coordinates), distance to the center of the goal, angle to goal, and body part (foot, head or other). Appendix A1 provides a detailed description of the model. Second, using the same dataset, we overlay a 1 by 1-meter grid on the field and compute the proportion of shots that originate from each grid cell. We also record the proportion of shots within a cell that are attempted by each body part. Third, we sample the location of $n$ shots from this distribution with $n \in \{50, 75, 100, 125, 150\}$. For each sampled shot, we derive the features and compute the shot's xG value, assuming that the player is an $\alpha$% better finisher: $xG = (1+\alpha/100) \times xG$, with $\alpha \in \{0, 5, 10, 15, 25\}$. The upper bound of 25% is selected based on Messi's aforementioned overperformance. We then sample an observed outcome (i.e., goal / no goal) from this modified distribution which we sum with the raw (i.e., unmodified) xG to obtain GAX. We repeat the procedure 10,000 times for robustness for each value of $n$ and $\alpha$.

Figure 2 shows a heatmap of the probability that a player's total number of goals will exceed their cumulative xG for each combination of skill level and shot volume. These probabilities indicate that while, in any given season, a player has a reasonable chance to overperform their xG, it would be hard to do this continually. For example, consider a player with $\alpha$ = 25 that takes 100 shots in each of five consecutive seasons. For this high volume and highly skilled finisher, the chances of overperforming cumulatively xG in at least four out of five seasons is 70.0% according to the binomial distribution. If the skill decreases to $\alpha$ = 10 even with upping the shot volume to 125 shots this probability drops to 41.6%.

[3] https://www.espn.com/soccer/insider/insider/story/_/id/38751980/predicting-el-clasico-analyzing-barcelona-real-madrid

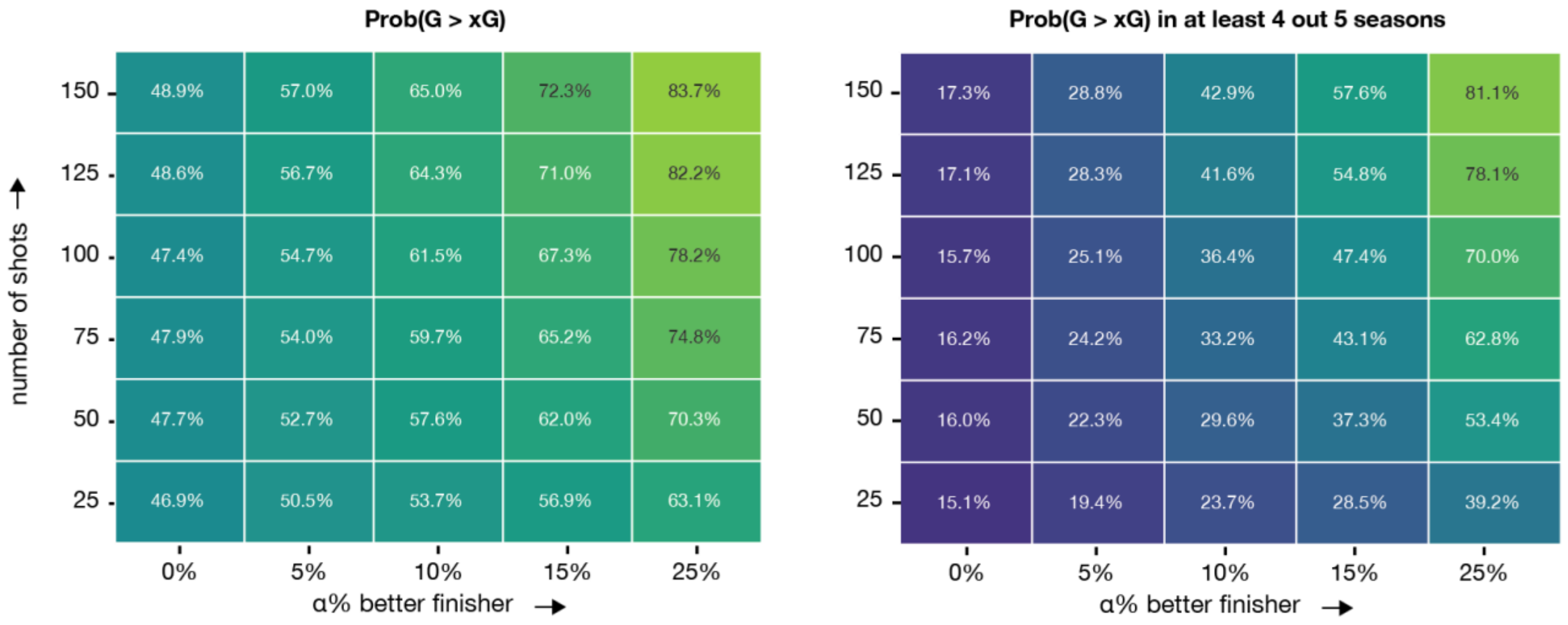

*Figure 2. A heatmap of the probability that a player's total number of goals will exceed their cumulative xG in a single season (left) and four out of five consecutive seasons (right) for each combination of skill level and shot volume per season. Players are unlikely to overperform their cumulative xG consistently unless they can pair exceptional finishing with high shot volumes.*

Table 1 shows the average and standard deviation of GAX in the simulations. Even for high skill levels and shot volumes, the averages are reasonably close to zero and there is a lot of variance. Again, this reinforces the point that you are unlikely to observe prolonged overperformance of cumulative xG unless a player pairs exceptional finishing with high volumes.

*Table 1. The average and standard deviation of GAX computed over 10,000 repetitions in the simulation for Hypothesis 1.*

| | **Number of shots** | | | | | |
|---|---|---|---|---|---|---|
| **Skill level** | **25** | **50** | **75** | **100** | **125** | **150** |
| **0%** | -0.00±1.39 | -0.02±1.97 | -0.01±2.42 | -0.06±2.78 | -0.00±3.06 | 0.01±3.43 |
| **5%** | 0.12±1.41 | 0.22±2.01 | 0.36±2.47 | 0.44±2.83 | 0.61±3.11 | 0.75±3.50 |
| **10%** | 0.23±1.43 | 0.48±2.03 | 0.73±2.52 | 0.93±2.87 | 1.23±3.16 | 1.48±3.55 |
| **15%** | 0.35±1.46 | 0.72±2.07 | 1.11±2.57 | 1.42±2.92 | 1.84±3.23 | 2.21±3.61 |
| **25%** | 0.60±1.51 | 1.22±2.14 | 1.85±2.65 | 2.38±3.00 | 3.08±3.34 | 3.70±3.73 |

The previous charts are based on an aggregate of all players. It is plausible that good finishers may shoot from different positions, affecting their ability to outperform xG. Therefore, we repeat the same experiment, except this time we compute the proportion of shots originating from each grid cell based on an individual player's level. Specifically, we use five seasons of shot data for 12 players:

Benzema, De Bruyne, Griezmann, Insigne, Kane, Lewandowski, Lukaku, Mbappé, Messi, Pogba, Salah, and Son. Figure 3 presents the results from this experiment as the difference in probability that a specific player's total number of goals will exceed their cumulative xG in a single season compared to the average player (i.e., the results in Figure 2). All these players take shots from locations that increase their odds of outperforming their cumulative xG compared to the generic shot profile that was previously considered. This indicates that overperformance is also tied to where a player shoots from.

### Hypothesis 2: Including all shots when computing GAX is incorrect and obscures finishing ability.

First, it may not be appropriate to include both headed and footed shots, as some players may be good with one body part and not the other. Of course, subdividing the shots based on this distinction could dramatically reduce the sample size, which may impact the statistical reliability of the analysis. Second, including speculative shots, such as long-range attempts near the end of a game when a team is trying to alter the outcome, can obscure the relationship. That is, these are shots that a player would not typically attempt but is only doing so due to game state factors. This is the analog to buzzer beaters or desperation shots when the shot clock is expiring in basketball or a Hail Mary in American football. However, their effect could be more pronounced in soccer due to the small number of shots taken by players in a season. Third, it may be warranted to exclude deflected shots from the analysis, as then another player has influenced the ball's trajectory. Hence, these shots cannot be used to measure a player's ability to accurately place the ball.

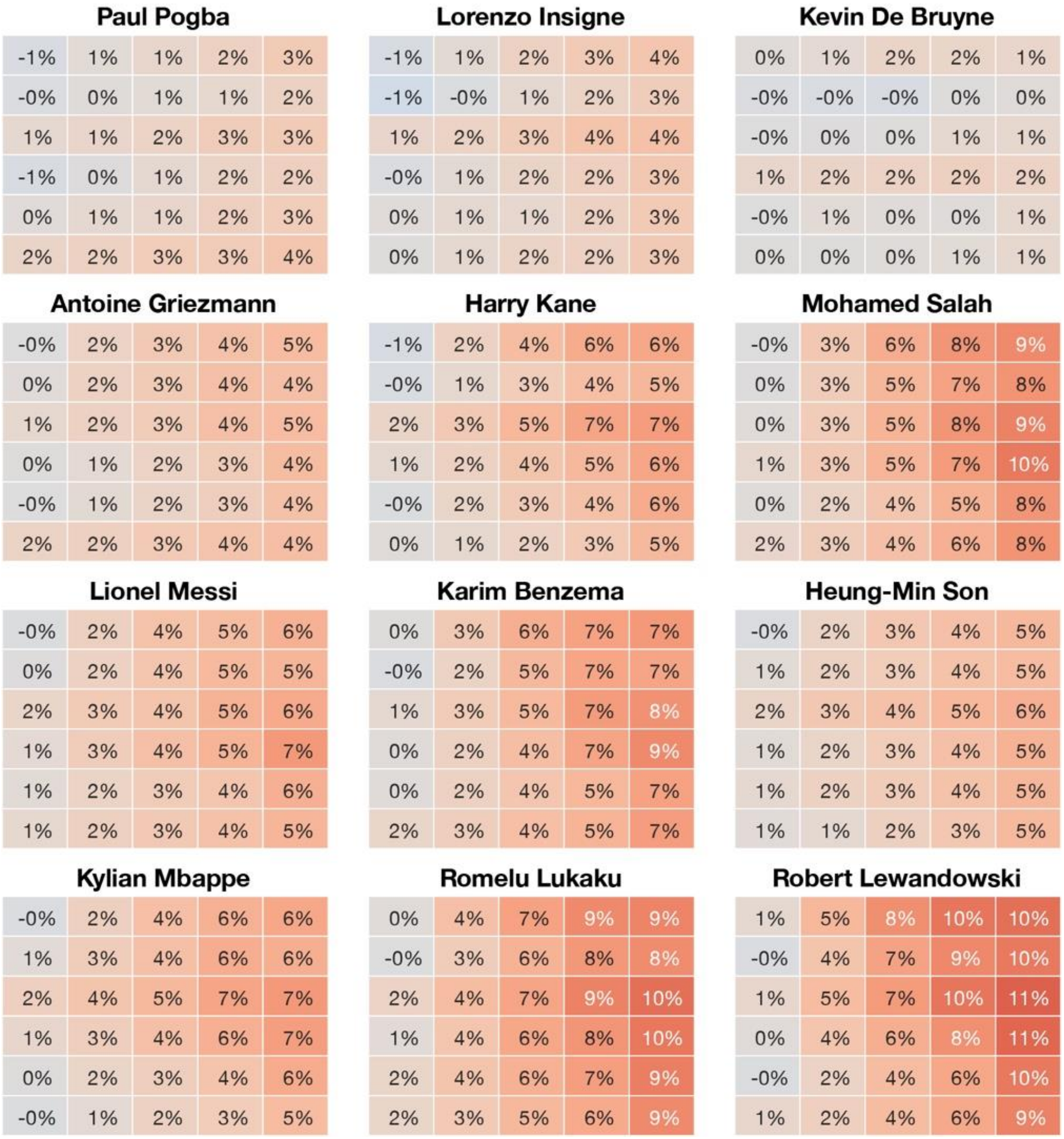

*Figure 3. The probability that some specific player's total number of goals will exceed their cumulative xG in a single season, visualized as the difference with the average player (Figure 2). Players who get high-quality scoring opportunities generally have a higher probability of outperforming their xG.*

To illustrate how the inclusion of all shots in the calculation of GAX can obscure an individual's finishing ability, we consider the 242 open play shots taken by Paul Pogba over the course of five Premier League seasons (2017/18 – 2021/22). Figure 3 shows the number of goals that an average player would realistically have scored from Pogba's shots, based on the Poisson binomial distribution [11]. Despite having a cumulative xG value of 20.85, Pogba only managed to score 17 goals. This suggests that he performs below the average level as a finisher. This contradicts the general perception of Pogba as an exceptional long-distance shooter.

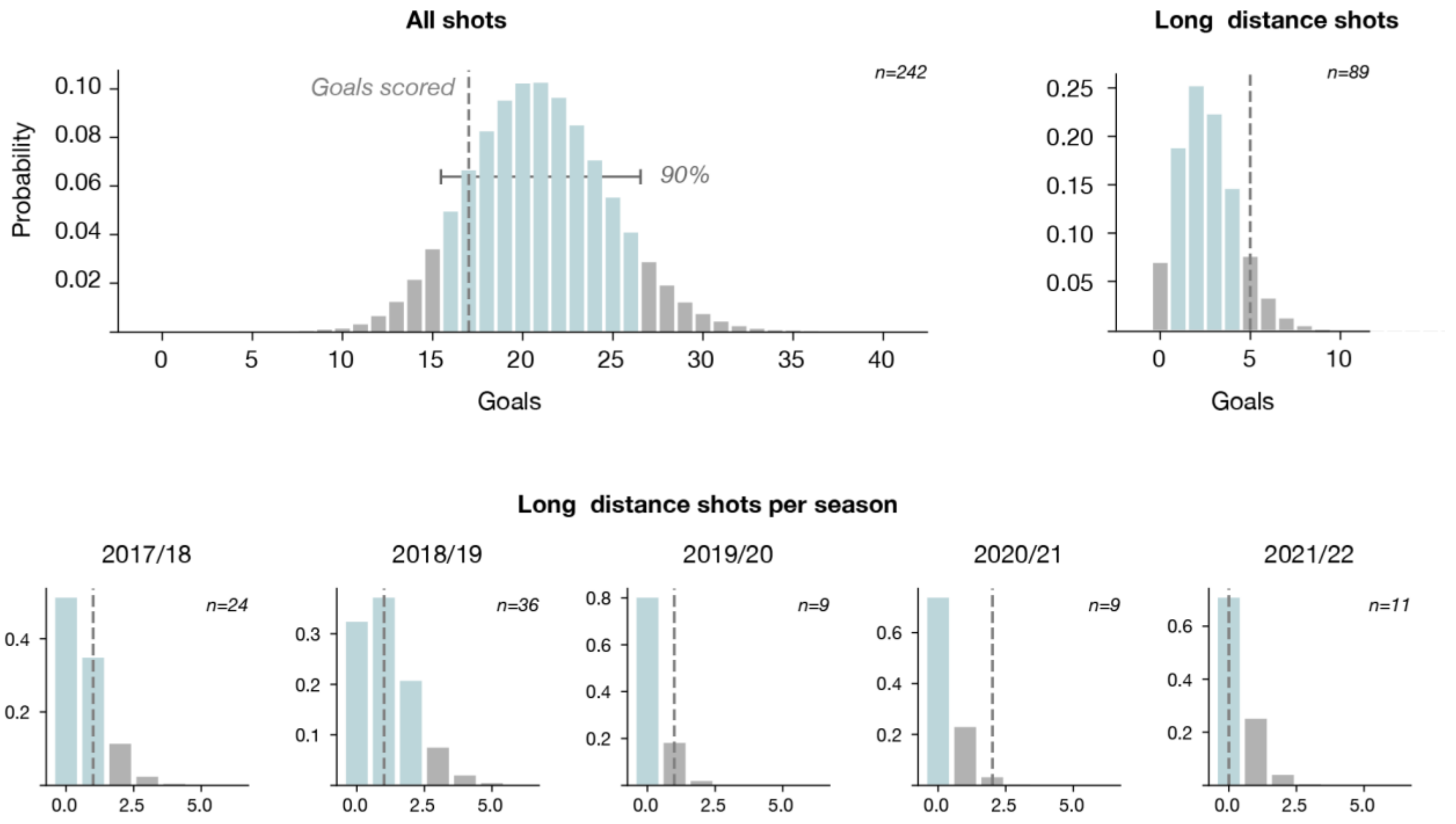

*Figure 4. Probability distributions for the number of goals scored, given the xG values of Paul Pogba's shots in the 2017/18 – 2021/22 Premier League seasons. Aggregating all shots, Pogba is overall a below-average finisher but excels at long-range shooting. The season-by-season sample size is insufficient to draw conclusions on Pogba's long-range shooting skill.*

Indeed, when focusing on long-distance shots (25 - 35 yards from goal), Pogba statistically outperforms his cumulative xG. However, when looking at these figures season-by-season, we can no longer draw conclusions on overperformance due to the limited sample sizes. Ultimately, drawing conclusions on over- or underperformance in specific shot categories is only feasible for exceptionally high-volume shooters over an extended period of time.

Overall, only 2.3% of shots are deflected. Nevertheless, they can have a significant effect on the figure of individual players, even those with high shot volumes. As an example, we consider the 289 shots taken by Riyad Mahrez during the 2017/18 – 2021/22 seasons, of which 17 were deflected (5.9%). With the inclusion of these deflected shots, Mahrez outperforms his xG by 14.61 goals; without the deflected shots, only by 9.03 goals.

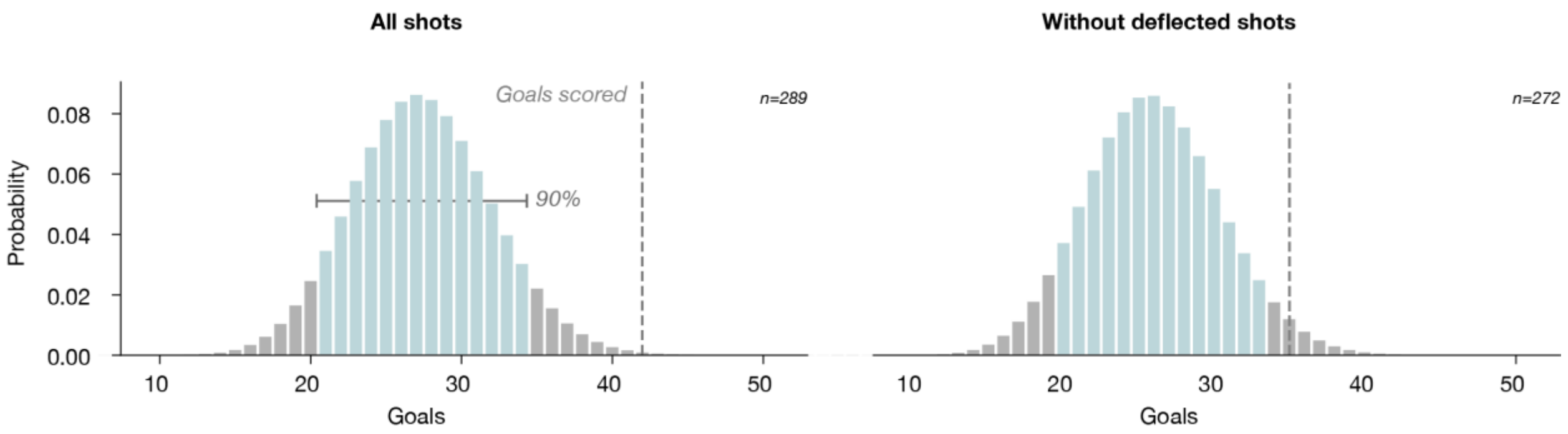

*Figure 5. Probability distributions for the number of goals scored, given the xG values of Riyad Mahrez's shots in the 2017/18 – 2021/22 Premier League seasons. When considering all of his shots, Mahrez demonstrates exceptional finishing abilities. Excluding the 17 deflected shots, his finishing skill is not regarded as exceedingly extraordinary.*

### Hypothesis 3: Interdependencies in the data bias GAX.

This bias arises because players have shots that appear both in the training and test data, creating a dependency between them. The effect will be most pronounced for players with large shot volumes. To illustrate the intuition for this reasoning, let's consider how xG models are trained and focus on the case of Lionel Messi.

Suppose the xG model is trained on all open play shots from Europe's top-5 leagues over the past five seasons, totaling around 250,000 shots. Within this dataset, Messi, known for his prolific shooting, contributes around 700 shots. Notably, Messi also consistently outperforms his cumulative xG. This could create a situation where Messi's strong performance influences the learned xG model, making the predicted xG values for shots similar to those he typically takes slightly higher compared to if his data were not included in the training set. This effect is analogous to the "good team bias" observed in possession value models [12].

The dependency between the training and test sets becomes problematic when applying the learned xG model to future shots taken by Messi. Omitting Messi's shots from the training data would yield a different model, leading to a higher GAX on his trademark finishes. This effect would likely extend to other players with a similar shot selection to Messi, causing their GAX values to also be slightly higher.

We perform two experiments to better understand the biases that arise from variations in the number of shots players take and their finishing skill on xG models. First, we explore how Messi's GAX varies based on the composition of the training dataset. We begin with the original xG model trained on the StatsBomb Open Data for the 2015/16 season (excluding shots by Messi) and compute Messi's GAX on StatsBomb's Messi Data Biography. We then augment the training set by generating between 0 and 5000 shots using the same procedure as for Hypothesis 1 but assign the label (i.e., goal/no goal) by sampling $n$ shots from distribution $(1+ \alpha/100) \times xG$ where xG comes from the model. We retrain the model after including these newly generated shots in the training dataset. We repeat the sampling and model training 100 times for each $\alpha$ and $n$ for robustness.

Figure 6 shows how Messi's GAX varies as we add more shots taken by above average finishers to the training data. As the training data becomes more biased towards containing shots from above average finishers, Messi's GAX declines. Adding shots from stronger finishers leads to bigger declines.

When we add 4000 shots from a finisher that performs 25% better than average, Messi's GAX drops from 127.6 to 120.8, a decrease by more than 5%. This experiment shows how the composition of the training data can have an effect on a player's GAX.

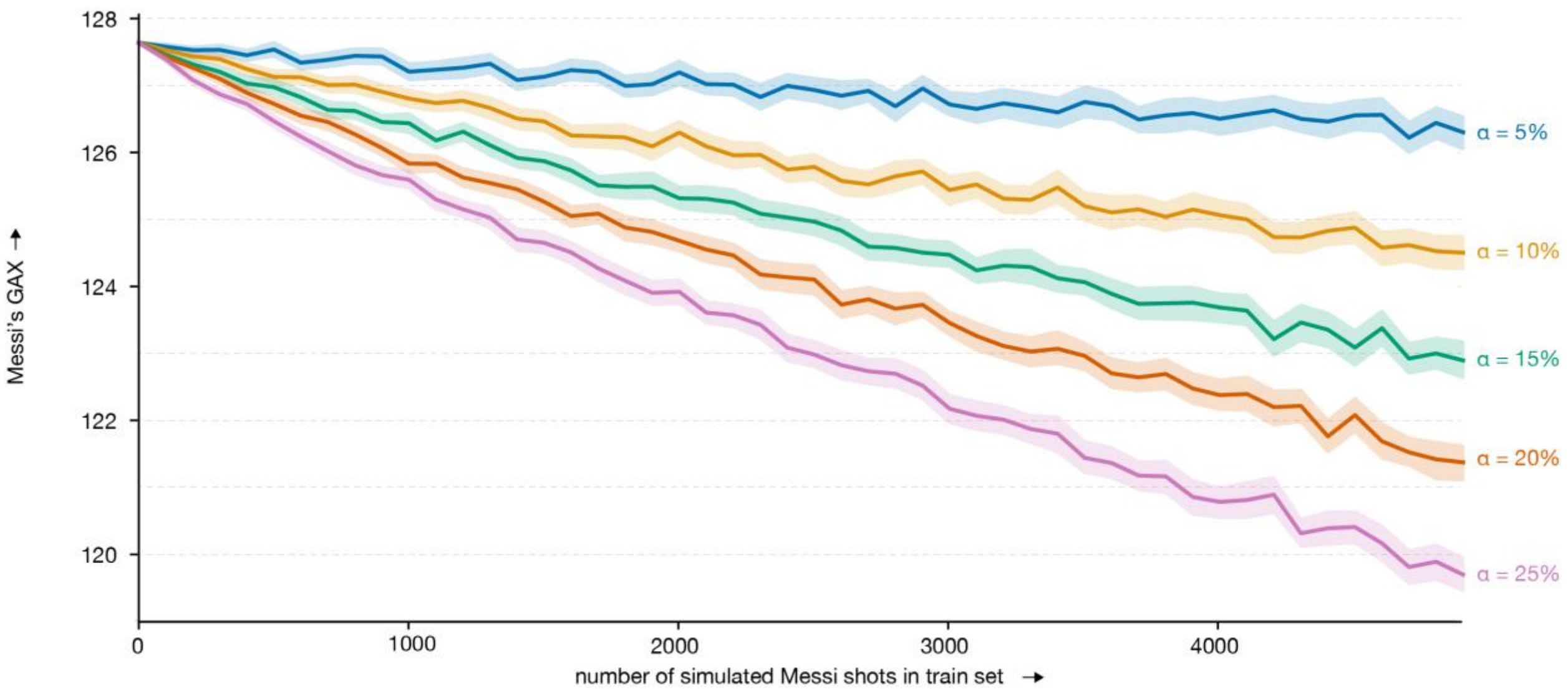

*Figure 6. Starting from an xG model trained on the StatsBomb Open Data for the 2015/16 season, an increasing number of simulated shots by above average finishers are added to the training set to illustrate how the composition of skill levels in the training data can have an effect on Messi's GAX. The shaded area represents the 95% confidence interval based on 100 runs.*

Second, we use a simulation study to better understand the biases that arise from variations in the number of shots players take and their finishing skill on xG models. We generate a training set of 1,000,000 shots using the same procedure as for Hypothesis 1 but assign the label (i.e., goal/no goal) by sampling from distribution $(1+ \alpha/100) \times xG$ where xG comes from the model and $\alpha \in \{-5, 0, 10, 20\}$. We consider three different allocations of the number of shots taken by players with each skill level: $\{(100k, 800k, 50k, 50k), (50k, 750k, 100k, 100k), (50k, 650k, 100k, 200k)\}$. We then train an xG model from each of these datasets.

Next, we estimate the effect of the skill distribution in the training data on the GAX of a player with $\alpha \in \{-5, 0, 10, 20\}$ better finishing skills who took $n \in \{75, 100, 125\}$ shots in a season. Therefore, we sample $n$ shots from distribution $(1+\alpha/100) \times xG$ in an identical manner and compute the xG value of these shots based on each model learned from the three training datasets, repeating it 10,000 times for each combination of $\alpha$ and $n$. For each sample, we compute the observed GAX and compare it to the average player GAX calculated by the ground truth model. As Figure 5 illustrates, omitting shots from excellent finishers from the training set leads to a higher GAX for these excellent finishers.

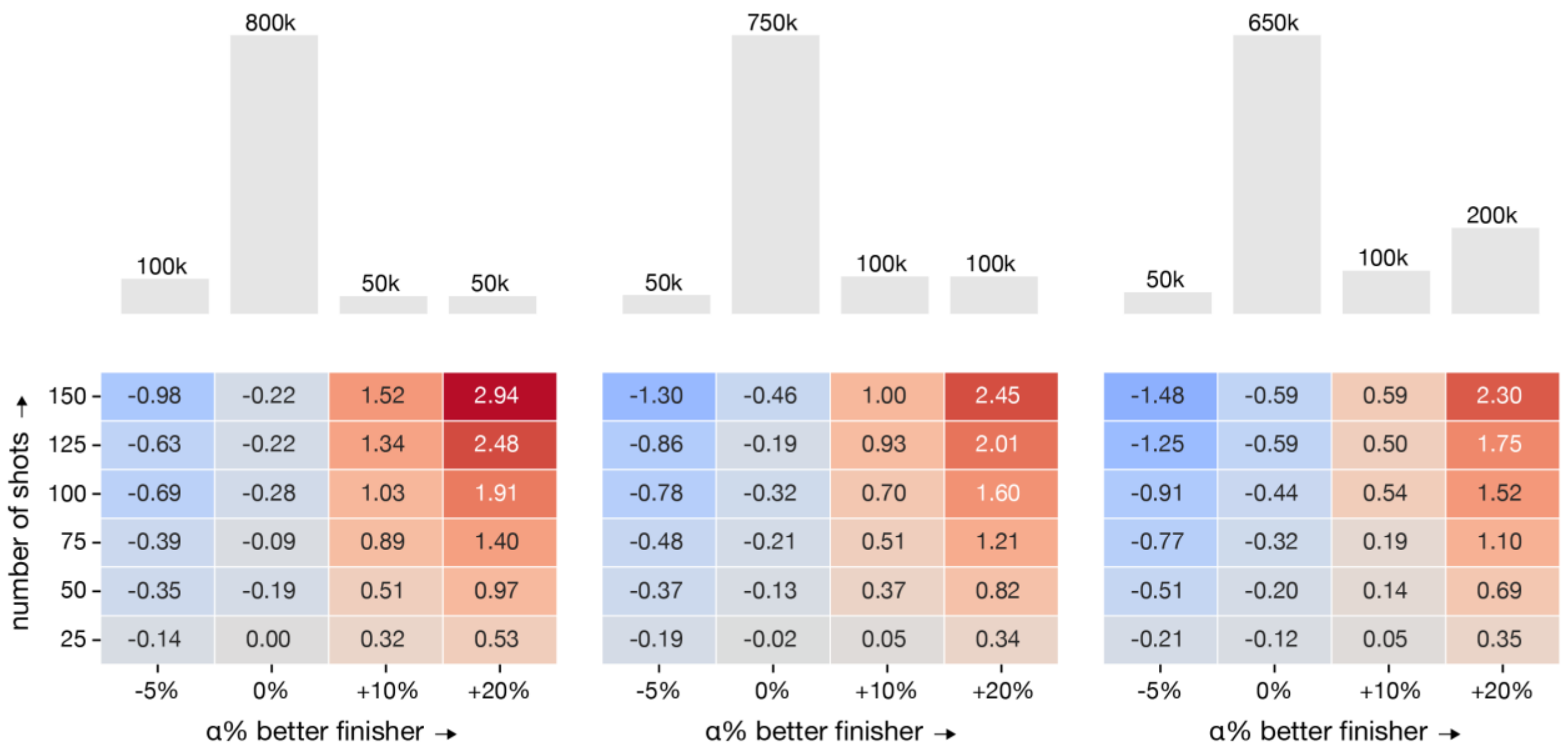

*Figure 7. The expected GAX for players with finishing skills α ∈ {−5, 0, 10, 20} and shot volumes n ∈ {25, 50, 75, 100, 125, 150}, in function of three training sets where the proportion of shots taken by players with each skill level is simulated according to the distribution displayed on top. When the training data contains more shots from excellent finishers, this leads to a decreased GAX for excellent finishers in the test set.*

# 3. Finishing Skill through the Lens of AI Fairness

The validity of GAX as a measure of finishing skill is based on the premise that xG captures the probability that the "average player" would finish a shot. However, a fundamental insight of this paper is that confounding effects in the training data mean that the model does not represent this "average player". As illustrated by the simulation experiment for Hypothesis 3, this bias creeps into the model, meaning that the model will overestimate the finishing skill of poor finishers and underestimate the skill of great finishers. The open question is how to solve this problem.

Our insight that xG models are biased yields an interesting parallel to the work on fairness in artificial intelligence.[4] This literature is concerned with understanding whether learned models systematically discriminate against certain groups, typically based on a protected attribute such as ethnicity or gender [13]. There are multiple definitions and concepts of fairness. At a high level, one type of metric tries to ensure that when a model's predictions are stratified by a protected attribute (i.e., partitioned by the values that the attribute can take on), the model performs similarly in each stratum. Concretely, predictive parity (also referred to as sufficiency) is particularly relevant for our goal of modeling the "average player" because it is tied to calibration which is how xG models are evaluated.

However, our goal differs from traditional predictive parity in two key aspects. First, traditional predictive parity requires that the scores of the model are calibrated by group. That is not what we want. Instead, among shots taken by players with a skill level α that are assigned probability $p$ of resulting in a goal, there should be $p + \alpha$ fraction of them that actually are converted. If this holds, the model is truly representative of the "average player". Thus, we do not strive for perfect calibration per group of players with a certain finishing skill, but instead want to underestimate the probability of scoring for above average finishers and overestimate the probability for below average finishers. For example, considering shots with an xG values of approximately 0.3 taken by players with skill level α = +10%, we aim for approximately 40% (30% + 10%) of those shots to result in goals.

[4] Mismeasuring finishing ability is clearly a trivial problem compared to the significant societal issues addressed in the fairness literature.

Second, our work differs from traditional fairness in that the protected attribute (i.e., a player's finishing skill level) is an unknown latent variable in our case. Therefore, we need to propose different properties that serve as a proxy for finishing skill in order to partition the shots into groups that highlight biases in the data. Based on our analysis and our limited sample of real shots, we consider three possible proxies:

1. Shot volume: Good shooters will take more shots. We divided players into three groups based on average shot volume per 90 minutes: low ($20^{th}$ percentile; < 0.875 shots per 90), medium, and high ($80^{th}$ percentile > 2.526). To deal with players that have limited playing time, we apply Laplace smoothing to the number of shots per 90 minutes, using as prior the median number of shots of an attacker (2.1 shots per 90), midfielder (1.1) or defender (0.4) depending on the player's most frequent starting position. We give this prior a weight of 270 minutes (which is equivalent to three games). Thus, we define the shot volume per 90 of player *i* as:
$$\theta_i = \frac{|shots|_i / |minutes|_i + 3 \times \theta_{pos}}{|minutes|_i + 270} \times 90$$
where $|shots|_i$ is the number of open play shots by player $i$, $|minutes|_i$ is the number of minutes and $\theta_{pos}$ is the average number of shots per 90 for the player's position.
2. Playing position: Attackers may be developed or slotted into this position because it was determined they were good shooters. Based on the StatsBomb position, we consider three groups: defenders, midfielders and attackers.
3. Team strength: Better teams have better players. Moreover, given that good teams are (much) stronger than most opponents they will take more shots and it is possible that they are easier to convert due to the opponents having weaker players. Using Club Elo ratings,[5] we divided the teams into three groups: potential UEFA Champions League (UCL) winners (four teams with Elo > 1950), the remaining teams in the top 25, and all other teams.

Figure 8 shows the calibration curves for each group within each of our three proxies. We use bins of with 1% with Gaussian kernel density estimation on the distribution of predicted probabilities and conversion rates in each bin to obtain a smooth curve. Also, we do not consider calibration for bins that contain less than 100 shots. We make the following observations:

- Partitioning by volume shows some miscalibration. Low volume shooters underperform whereas high volume slightly overperform. Average volume is well-calibrated.

[5] http://clubelo.com/

- Partitioning by position shows that defenders are miscalibrated and underperform xG. Interestingly, midfielders and attackers are reasonably well calibrated by xG values < 0.25. There is some deviation above this, which can partially be ascribed to the smaller number of shots here.
- Team strength shows less miscalibration with all models being well calibrated for xG values < 0.1. There is some deviation for the potential UCL winner group but this is likely due to sample size issues. The other teams in the top 25 slightly outperform xG for shots valued between 0.1 and 0.2.

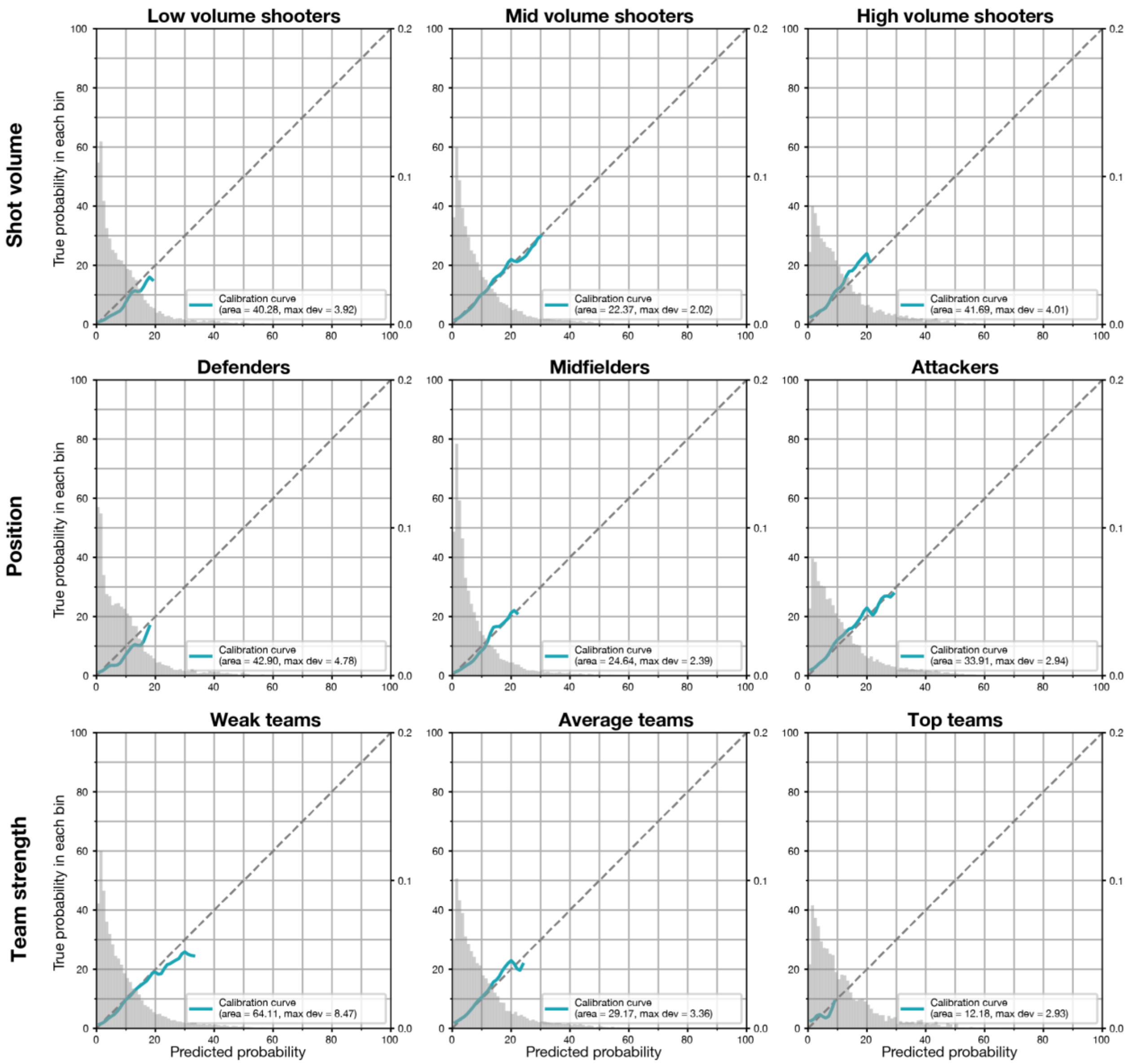

*Figure 8. Probability calibration curves for shots grouped by shot volume, playing position and team strength. We only evaluate the calibration in xG ranges with sufficient shots (n > 100). The histograms display the distribution of xG values within each group.*

The calibration curves support our hypothesis that low volume shooters, defenders and players of weaker teams are in general worse at finishing. Yet, this does not necessarily mean that the model accurately represents the "average player". The magnitude of the deviation might still be inaccurate. Therefore, we further investigate the effect of shot volume as a source of bias. Figure 9 shows the conversion rates in function of the distance to the center of the goal for both footed and headed open-play shots for each of low, mid and high-volume shooters. The distances were discretized into five ranges: 0-5m, 5-11m, 11-16m, 16-25m and >25m. For both types of shots, we see a clear ordering for every distance, with high volume shooters having better conversion rates than mid volume ones, and the mid volume shooters having better conversion than low volume shooters. This provides additional evidence that there is a link between shot volume and finishing skill. Moreover, the difference in conversion rates between low, mid and high-volume shooters seems larger than what is reflected in the calibration plots.

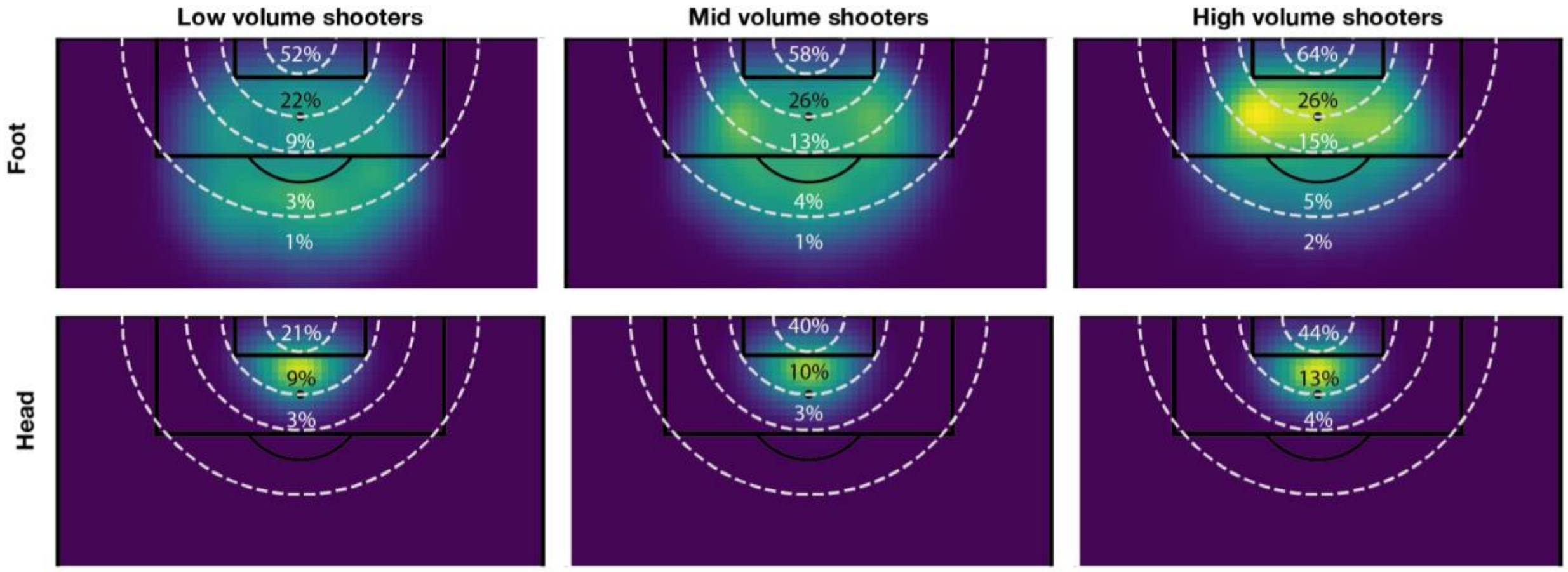

*Figure 9. The spatial distribution and conversion rates of shots by foot and headers, grouped by the shot volume of the player who takes the shot. The conversion rates were computed for five distance ranges: 0-5m, 5-11m, 11-16m, 16-25m and >25m.*

These analyses show that figuring out what an "average player" exactly is depends on how different characteristics work together, and the model needs to consider all these factors. To this end, we use the idea of multi-calibration which has arisen in the AI fairness literature [14]. This class of techniques attempts to calibrate a model's predictions for subpopulations that may be defined by complex intersections of many attributes. The power of multi-calibration lies in its ability to explicitly represent various player types based on their characteristics within a single model. Effectively, this provides multiple alternative representations of the average player based on different group characteristics. Consequently, we can explicitly state characteristics of the player group used to derive the GAX metric. This approach contrasts with the typical xG model, where these characteristics and any associated biases remain implicit.

We apply multi-calibration as a post-processing step to our logistic regression xG model to obtain calibrated predictions within each of our subgroups defined by the player's shooting volume and playing position. This leads to nine baseline representations of an "average player". We do not consider team strength due to the small sample sizes in the “potential UCL winners” group. For further details about the training procedure and calibration curves, see appendix A2.

To illustrate how multi-calibration works, we consider the Messi biography data provided by StatsBomb. In this dataset, Messi scored 375 goals from 1862 open play shots. The standard logistic regression xG model that we trained assigns these shots a cumulative xG of 247.43. According to the StatsBomb xG values available in the Messi biography, these shots have a cumulative xG of 246.00. Figure 10 shows the cumulative xG of Messi's shots after applying multi-calibration for each of the nine "average players." If we compare him to the average high-volume attacker, he would have a cumulative xG of 274.3. Remarkably, even compared to this elite group that contains excellent finishers, he still substantially outperforms his xG.  We can also weigh each group of the proportion of players in that group to create a "weighted average player.” Compared to this baseline, Messi would have a cumulative xG of 225.01 meaning his GAX has increased from 127.57 to 149.99 which is around a 17% increase. This gives an indication about how the overrepresentation of shots from good finishers in the data used to train the baseline xG model can bias GAX in practice.

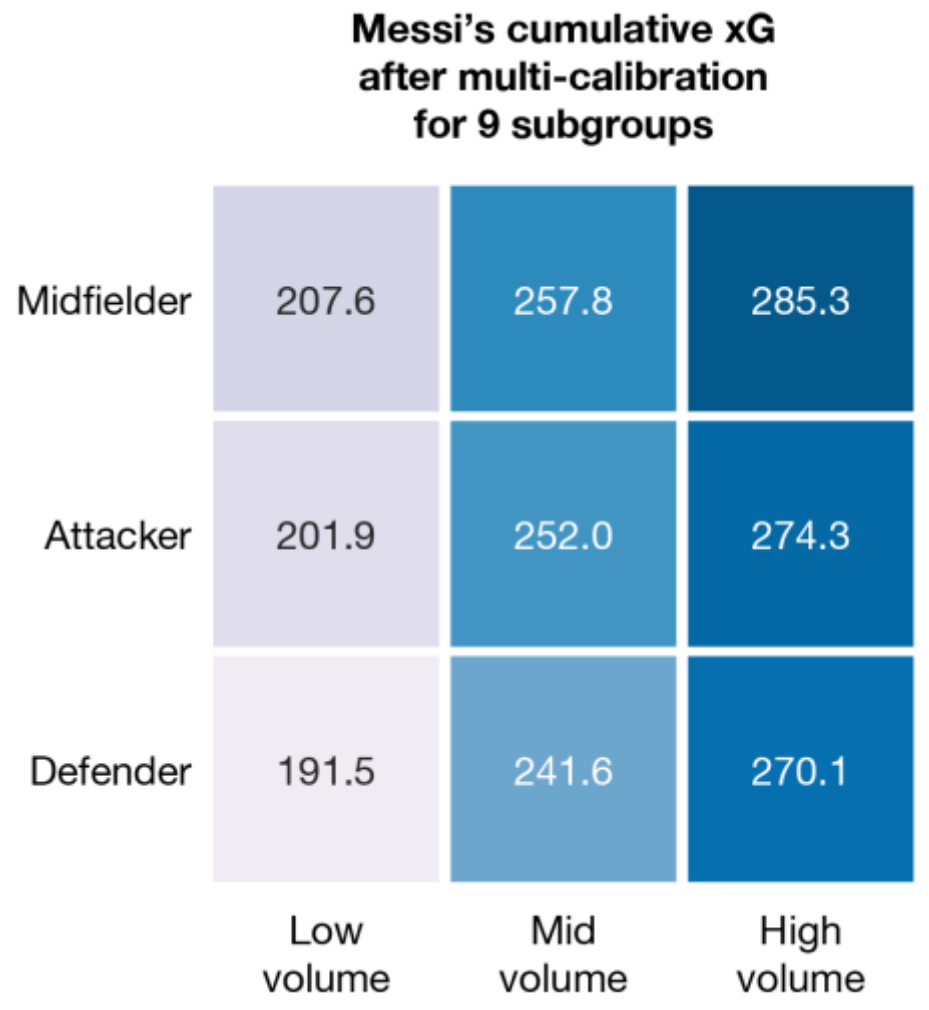

*Figure 10. The technique of multi-calibration allows comparing Messi’s performance against alternative notions of an “average player”, defined based on playing position and shot volume. With a total of 375 goals from 1862 open-play shots, Messi outperforms the expectations in each subgroup.*

We conducted an analogous analysis of the 2015/16 Premier League season, computing xG using the standard logistic regression model and the multi-calibrated model (Figure 11). We focused on players who scored at least five goals (n=63). Among them, 50 players exceeded their standard xG by an average of 16.72%, while 51 players outperformed their multi-calibration-based xG by an average of 20.00%. Only 47 players surpassed their StatsBomb xG, showcasing an average outperformance of 12.78%. These findings show that the standard xG models might underestimate the finishing skill of good finishers (the selection of players that scored at least five goals is naturally biased towards better finishers). Moreover, the complex StatsBomb xG model might pick up more of the bias in the data. In contrast to the absolute GAX values, the induced ranking of players is not significantly affected. Appendix A3 contains a table with the raw values for the top 25 players.

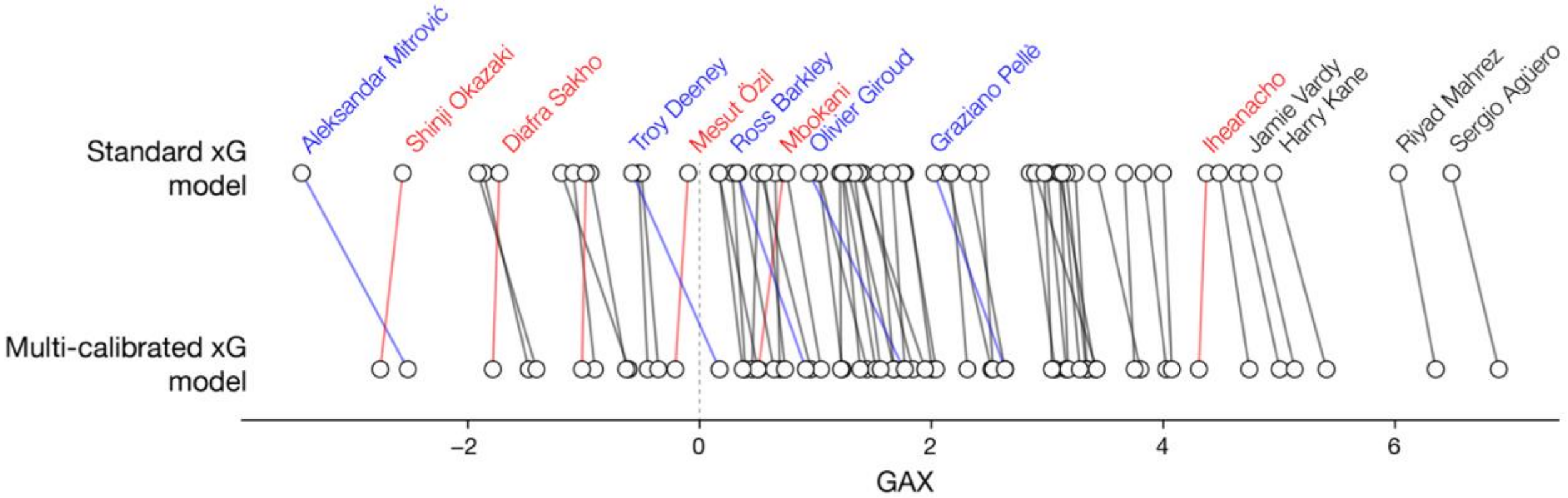

*Figure 11. Contrasting GAX derived from the standard logistic regression model and the multi-calibrated xG model for players in the 2015/16 Premier Leagues season who scored at least five goals. Players experiencing the most significant GAX increases and decreases compared to the standard model are highlighted in blue and red, respectively.*

# 4. Related Work

In the pre-xG era, evaluating a player's finishing skill often relied on conversion rate, that is, the proportion of shots that a player takes that result in a goal. Intuitively, higher conversion rates indicate better finishing ability. Gregory [15] posits that this is biased because good shooters take more shots which affects the overall average. He overcomes this by using an empirical Bayesian approach. However, this work does not account for shot quality. Similarly, Shaw [16] proposes using empirical Bayes for adjusting GAX (albeit defined as the ratio G/xG) for the variation in volume and quality among players.

Kwiatkowski [17] looked at assessing finishing skill by employing Bayesian logistic regression. This work included an indicator variable for each player in the feature set of the xG model. The sign and magnitude of the weight learned for each player are indicative of a player's finishing ability. Moreover, the Bayesian approach gives a measure of uncertainty to these parameters and if the range of likely values does not include zero then this gives a stronger belief about their estimated skill. Arslan [18] took a similar Bayesian approach but focused on comparing a generic model to a model fine-tuned to Messi, given that he is widely considered to be an excellent finisher.

It has also been noted that the cumulative xG is a point estimate of a distribution over possible observed goal counts. Therefore, how a player accumulates xG is important [11]. This has led to the proposal of an alternative measure for finishing ability which is to compute the chance of observing the actual number of goals scored given the xG values for the shots that were taken via a Monte Carlo simulation [19].

Finally, arguably finishing skill involves more than just placing the ball on target; it also encompasses the ability to place the ball accurately and strategically in locations that are challenging for the goalie to reach. This can be captured via post-shot xG models [20], [21]. However, these treat missing just wide of the goal the same as shooting the ball nowhere near the goal and accounting for how close a player is to the goal when they miss can be beneficial [22]. Baron et al. [23] leveraged this insight to propose a new set of finishing skill metrics that attach non-zero values to off-target shots. Effectively, this increases the sample sizes. Still, it is important to note that these models are still trained from the same data used to train xG models. Hence, they will still suffer from the data biases we described arising from the overrepresentation of shots from certain groups of players. The idea of multi-calibration is also applicable to these analyses.

# 5. Discussion

The assessment of skill in soccer commonly entails comparing a player's performance to that of the "average" or "typical" player. This is where expected value metrics, like xG, play a crucial role by estimating how an average player would perform in a similar situation.

However, these metrics rely on models that are trained to produce calibrated probabilities, aiming to faithfully represent the observed data. Nonetheless, the observed data are likely subject to biases, such as the proportion of shots from good finishers being higher. This limits the metrics' representativeness of a typical player's performance. Effectively, the model is more of a weighted average where more frequent shooters receive more weight and hence have a large effect. As an illustrative example, our simulations showed that the presence of above-average finishers in the model's training data may result in a small but consistent underestimation of the abilities of strong finishers. As a result, a fundamental tension arises between two objectives. While faithfully modeling xG values for a typical player would aid in skill assessment, it may not provide as accurate values for the empirically observed distribution of shots. Conversely, prioritizing the estimation of accurate probability estimates diminishes their usefulness for assessing skill.

Effectively modeling xG values for a typical player poses challenges due to the inherent ambiguity in defining what precisely constitutes "average." In this paper, we propose using the AI fairness technique of multi-calibration to explicitly define various notions of an "average player" by learning a model that is calibrated within various subgroups of players with specific characteristics. This approach enables practitioners to assess under- or overperformance with respect to distinct, well-defined categories of players. For example, players can be evaluated with respect to their playing position or role, anthropometric characteristics, and level of competition in which they operate.

Nevertheless, even with an accurate representation of the "average" player, drawing conclusions on over- or underperformance on finishing skill seems to be possible only in a few occasions. The deviation between the actual and expected results is often moderate compared to the variance such that even players with average finishing skills have a reasonable likelihood of outperforming their xG. Unless an individual is both a high-volume shooter and an exceptionally proficient finisher, sustained overperformance is unlikely.

Finally, the inclusion of all shots in GAX can obscure the true finishing ability of players, as demonstrated by the case of Pogba and Mahrez. First, finishing skill encompasses various facets related to footedness, heading and long-range finishing, each of which should ideally be assessed separately. However, subdividing the shots into even smaller samples would result in wider confidence intervals, limiting further the conclusiveness of the analysis. Second, not all shots can accurately reflect a player's finishing skill. For instance, deflected shots and speculative attempts may not provide a reliable measure. Isolating and excluding such shots can pose a challenge.

From a practical point of view, we believe that this paper raises a number of points that people training and using models such as an xG model should consider:

1. It is important to consider the end goal of the analysis. If the objective is to have a high-level understanding of what are good places to shoot from and which players tend to take promising shots, then the standard xG analysis is likely appropriate. If the objective is to investigate skill, the crucial question is to carefully think about the definition of an average player and weigh the data in such a way that you learn a model of an average player. Finally, gaining insight into what shots a specific player should take would require incorporating player information into the model.
2. The composition of the data set can have a (large) effect on the learned model and subsequent analysis. Our experiments showed that adding more data from skilled players results in underestimates for good finishers. More generally, assuming that finishing skill is not evenly distributed across leagues, training an xG model on data just on the top 5 leagues versus the top 25 leagues will likely influence a metric such as GAX. Therefore, knowing what data was used to train the model can help contextualize its results.
3. It is important to consider what actions are included and excluded from the analysis. In the context of finishing skill, the typical computation of GAX may include deflected shots. As seen, this had a large effect on Mahrez's GAX. More generally, rare or atypical shots (e.g., Patrik Schick's lob versus Scotland at EURO 2020) may skew results because their probabilities may not be accurately modeled and converting them, particularly if they have a low probability, will result in a large positive contribution to GAX. Therefore, it would be beneficial to explicitly state what is (not) included and why.
4. How much of the bias in the data is picked up by a model will likely depend on the considered features and the model type. Including more complex features and using an expressive model class such as gradient-boosted trees as is done with most deployed xG models would likely result in more bias in the estimates due to their ability to better fit the observed data. This is important to investigate.
5. The AI fairness literature provides tools for better understanding a model's biases and how it performs on different groups within the data. Using these ideas can help people better quantify the performance of learned models and when they are and are not applicable.

In summary, while GAX can serve as a valuable metric, it is essential to acknowledge the inherent variability of finishing skill, consider the selection of shots to analyze, and account for potential data biases.

**Acknowledgements.** This research received funding from the Flemish Government (AI Research Program), the KU Leuven Research Fund (iBOF/21/075), and FWO-Vlaanderen (K801724N) to Jesse Davis. We thank Jan Van Haaren for extremely valuable feedback on the manuscript and for discussing the ideas with us. We thank StatsBomb for generously providing the data used in this research.

StatsBomb

# Appendix

## A1. Standard Logistic Regression xG Model

We train a basic logistic regression xG model on public event stream data from StatsBomb. Our dataset comprises 43,110 open play shots (excluding own goals) from the Big 5 European leagues in the 2015/16 season. We randomly split them into a training and test set with a 90 : 10 distribution. We applied a stratified split, so the successful/missed shot ratio remains the same across training and test sets. The test set is left as hold-out data, and results on performance are reported for this dataset.

As features, we use the start and end coordinates of the shot, the distance and angle from the goal, and a dummy encoding of the body part used to take the shot. These are the key features in any xG model. Additional contextual features result in marginally better estimates of expected goals, but these would make the sampling experiments too complicated.

We build the model using Scikit-Learn's logistic regression implementation with the Newton-Cholesky solver and L2-regularization (C=1). Table 2 shows the model's coefficient values after training. We obtain an AUROC-value of 0.7990 and a Brier score of 0.0793.

*Table 2. Features and their corresponding weight in the standard logistic regression xG model.*

| FEATURE | DESCRIPTION | WEIGHT |
|---|---|---|
| start_x | x-coordinate of the shot | -0.12903395599643944 |
| start_y | y-coordinate of the shot | +0.0007081390917350903 |
| distance | Distance to the center of the goal (in meters) | -0.31351026346703825 |
| angle | Angle to the center of the goal (in radians) | +0.09095528657471205 |
| bodypart_head | True if the shot was a header | -1.2946488935455573 |
| bodypart_other | True if the shot was taken with a bodypart other than the foot or head | -0.19292432746094432 |
| (intercept) | | +14.301040398979099 |

## A2. Multi-Calibrated xG Model

To obtain a multi-calibrated model, we post-process the standard logistic regression-based xG model (see appendix A1) with *pmcboost*.[6] We asses multi-calibration over subgroups defined by playing position and shot volume, using a calibration error tolerance of 0.01 per group with 10 log-spaced calibration bins [0.00, 0.015, 0.023, 0.034, 0.052, 0.079, 0.12, 0.18, 0.27, 0.40, 1.00] and limiting the number of iterations to 100 to avoid overfitting. Figure 11 shows the resulting calibration curves. The model is clearly better calibrated in each subgroup than the standard model (see Figure 8).

[6] Implementation is available at http://github.com/cavalab/pmcboost.

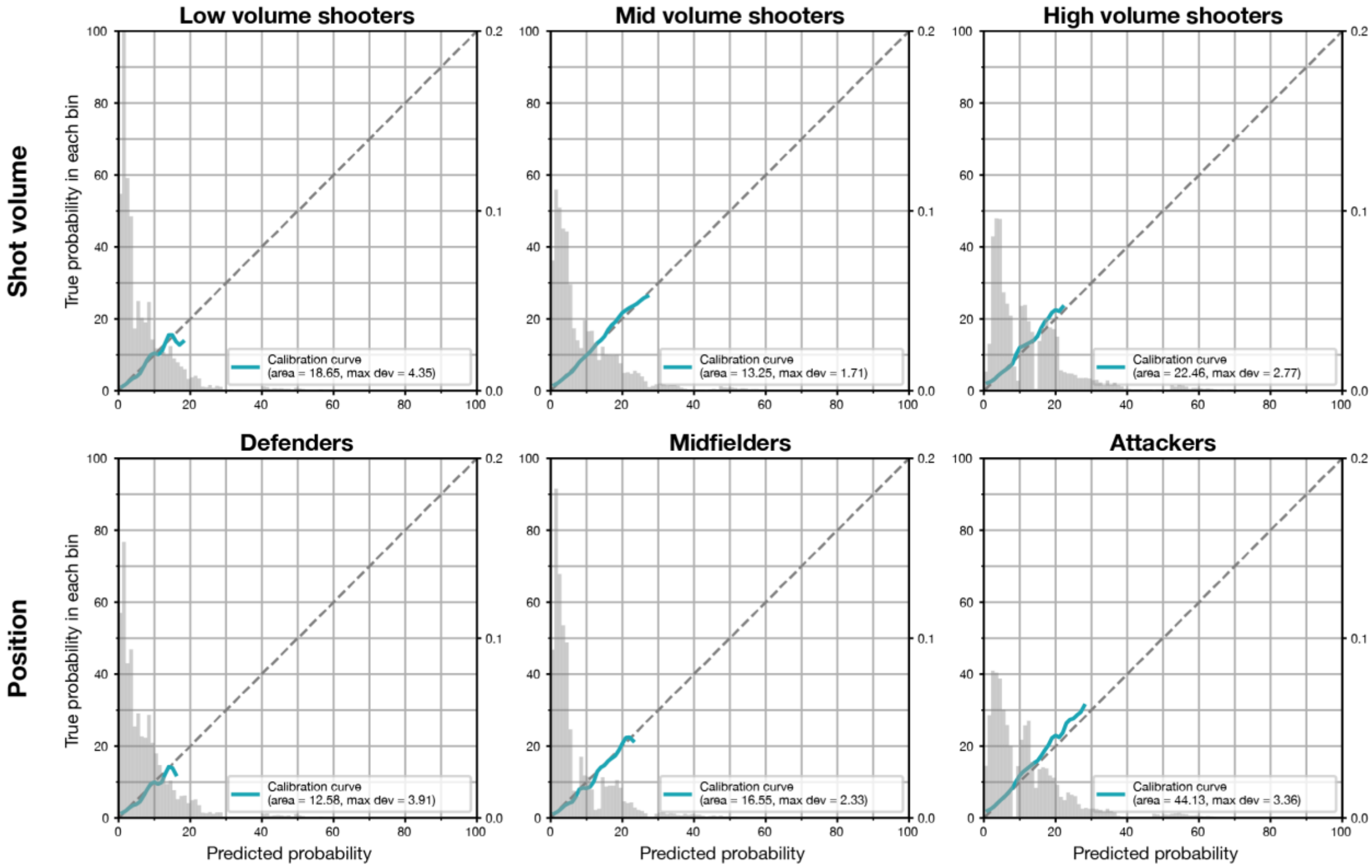

*Figure 12. Probability calibration curves for shots grouped by shot volume, playing position and team strength after post-processing the model with multi-calibration. We only evaluate the calibration in xG ranges with sufficient shots (n > 100). The histograms display the distribution of xG values within each group.*

## A3. Contrasting xG Values Derived from Biased and Unbiased Models

Table 3 compares the GAX derived from StatsBomb's xG values, the standard xG model and the multi-calibrated model for players that scored at least five goals in the 2015/16 Premier League season.

*Table 3. Comparison between the GAX based on the StatsBomb xG values, standard logistic regression xG model and the multi-calibrated xG model. The top 25 players with the highest GAX in the 2015/16 Premier League season according to the standard xG model are included. Players that scored less than five goals are excluded.*

| | PLAYER | GOALS | STATSBOMB | | STANDARD | | MULTI-CALIBRATED | |
|---|---|---|---|---|---|---|---|---|
| | | | xG | GAX | xG | GAX | xG | GAX |
| 1 | Sergio Agüero | 20 | 13.71 | 6.29 | 13.51 | 6.49 | 13.10 | 6.90 |
| 2 | Riyad Mahrez | 13 | 7.16 | 5.84 | 6.97 | 6.03 | 6.65 | 6.35 |
| 3 | Harry Kane | 20 | 17.71 | 2.29 | 15.05 | 4.95 | 14.59 | 5.41 |
| 4 | Jamie Vardy | 19 | 17.55 | 1.45 | 14.26 | 4.74 | 13.87 | 5.13 |
| 5 | Anthony Martial | 11 | 6.90 | 4.10 | 6.36 | 4.64 | 6.00 | 5.00 |
| 6 | Roberto Firmino | 10 | 7.30 | 2.70 | 5.51 | 4.49 | 5.26 | 4.74 |
| 7 | Iheanacho | 8 | 5.05 | 2.95 | 3.63 | 4.37 | 3.69 | 4.31 |
| 8 | Georginio Wijnaldum | 10 | 6.79 | 3.21 | 6.00 | 4.00 | 5.93 | 4.07 |
| **9** | Daniel Sturridge | 8 | 5.06 | 2.94 | 4.17 | 3.83 | 3.97 | 4.03 |
| **10** | Kevin De Bruyne | 7 | 4.44 | 2.56 | 3.33 | 3.67 | 3.25 | 3.75 |
| **11** | Bamidele Alli | 10 | 8.40 | 1.60 | 6.57 | 3.43 | 6.20 | 3.80 |
| **12** | Aaron Lennon | 5 | 1.95 | 3.05 | 1.76 | 3.24 | 1.66 | 3.34 |
| **13** | Divock Origi | 5 | 2.31 | 2.69 | 1.83 | 3.17 | 1.67 | 3.33 |
| **14** | Nathan Redmond | 6 | 2.56 | 3.44 | 2.87 | 3.13 | 2.72 | 3.28 |
| **15** | Marcus Rashford | 5 | 1.80 | 3.20 | 1.89 | 3.11 | 1.82 | 3.18 |
| **16** | Dimitri Payet | 7 | 4.37 | 2.63 | 3.90 | 3.10 | 3.59 | 3.41 |
| **17** | Jermain Defoe | 14 | 12.43 | 1.57 | 11.00 | 3.00 | 10.84 | 3.16 |
| **18** | Steven Davis | 5 | 2.06 | 2.94 | 2.03 | 2.97 | 1.96 | 3.04 |
| **19** | Shane Long | 10 | 7.98 | 2.02 | 7.11 | 2.89 | 6.57 | 3.43 |
| **20** | André Ayew | 11 | 7.54 | 3.46 | 8.15 | 2.85 | 7.93 | 3.07 |
| **21** | Bojan Krkíc | 5 | 3.19 | 1.81 | 2.58 | 2.42 | 2.47 | 2.53 |
| **22** | Pedro | 7 | 5.98 | 1.02 | 4.68 | 2.32 | 4.37 | 2.63 |
| **23** | Manuel Lanzini | 5 | 3.34 | 1.66 | 2.83 | 2.17 | 2.69 | 2.31 |
| **24** | Ayoze Pérez | 6 | 3.69 | 2.31 | 3.86 | 2.14 | 3.48 | 2.52 |
| **25** | Graziano Pellè | 11 | 7.72 | 3.28 | 8.98 | 2.02 | 8.36 | 2.64 |